\documentclass{article}

\usepackage{arxiv}

\usepackage[utf8]{inputenc} 
\usepackage[T1]{fontenc}    
\usepackage{url}            
\usepackage{booktabs}       
\usepackage{amsfonts}       
\usepackage{nicefrac}       
\usepackage{microtype}      %
\usepackage{lipsum}		
\usepackage{amsmath}
\usepackage{amssymb}
\usepackage{amsthm}
\usepackage{graphicx}
\usepackage{float}
\usepackage[normalem]{ulem}
\useunder{\uline}{\ul}{}
\usepackage{subfig}
\usepackage[colorlinks,linkcolor=blue]{hyperref}

\usepackage{listings}
\usepackage{thmtools}
\usepackage{hyperref}

\usepackage{tikz}
\usepackage{titlesec}
\usepackage{fullpage}
\usepackage{lastpage}
\numberwithin{figure}{section}
\title{A Method to Model Conditional Distributions with Normalizing Flows}

\author{
  Zhisheng Xiao \\
  Computational and Applied Mathematics\\
  University of Chicago\\
  Chicago, IL, 60637\\
  \texttt{zxiao@uchicago.edu} \\
   \And
  Qing Yan \\
  Department of Statistics\\
  University of Chicago\\
  Chicago, IL, 60637\\
  \texttt{yanq@uchicago.edu} \\
    \And
  Yai Amit \\
  Department of Statistics\\
  University of Chicago\\
  Chicago, IL, 60637\\
  \texttt{amit@marx.uchicago.edu} \\
}

\begin{document}
\maketitle

\begin{abstract}
In this work, we investigate the use of  normalizing flows to model conditional distributions. In particular, we use our proposed method to analyze inverse problems with invertible neural networks by maximizing the posterior likelihood. Our method uses only a single loss and is easy to train. This is an improvement on the previous method that solves similar inverse problems with invertible neural networks but which involves a combination of several loss terms with ad-hoc weighting. In addition, our method provides a natural framework to incorporate conditioning in normalizing flows, and therefore, we can train an invertible network to perform conditional generation. We analyze our method and perform a careful comparison with previous approaches. Simple experiments show the effectiveness of our method, and more comprehensive experimental evaluations are undergoing. 
\end{abstract}

\section{Introduction}\label{intro}

Deep generative models are powerful tools to model complicated distributions. They either model the distributions implicitly without assuming any parametric specifications, or explicitly with some parametrized form. The implicit models are dominated by Generative Adversarial Networks (GANs)\cite{GAN}, while explicit models have many variants, including Variational Auto-encoders (VAEs)\cite{VAE1, VAE2}, Normalizing Flows\cite{NICE, RNVP,GLOW2}, and auto-regressive models\cite{PIXEL, MASK}. They have been shown to successfully model very complicated distributions, such as high resolution images\cite{Biggan, vqvae, GLOW2} and produce high quality samples from the learned distributions. 

While these deep generative models obtain great results in modeling the marginal distribution of data, many important tasks require models of conditional distributions. For example, image coloring\cite{Color1, Color2} requires an estimate and samples from the distribution of color channels conditional on the luminance channel. Similarly, image-to-image translation\cite{trans1, trans2} requires  samples from the distribution of images in domain $A$ conditional on an image in domain $B$. In physical science, inverse problems involving  estimation of the posterior distribution of some parameters given observations are ubiquitous. Therefore, it would be desirable to extend deep generative models, which are very successful in modeling marginal densities, to the task of modeling conditional distributions. Some attempts have been made to add conditioning to GANs or VAEs\cite{CGAN,CVAE}. However, similar extension based on normalizing flows have not attracted much attention.

Normalizing flows are invertible neural networks that continuously transform a simple density into the complicated data density. They have been shown to successfully  model a wide range of distributions. Recently, \cite{INN} proposed to use invertible neural networks to analyze inverse problems, which is a task involving the modeling of conditional distributions. They further introduced cINN\cite{CNF}, which is a general framework for incorporating conditioning in invertible neural networks. This work is motivated by their ideas, and we aim at finding an alternative way to model conditional distributions with normalizing flows. We compare our method with their approaches, and apply our methods to the task of analyzing inverse problems and conditional generation.

The rest of the paper is organized as follows. In Section \ref{bg}, we introduce the basics of normalizing flow and inverse problems, as well as review some related work. In Section \ref{our} and \ref{cnf}, we introduce our method for inverse problems and conditional generation. In Section \ref{exp}, we demonstrate the effectiveness of our model with some simple experiments. More comprehensive experimental evaluations are undergoing. 

\section{Backgrounds and Related Works} \label{bg}

\subsection{Invertible Neural Networks}

As the name suggested, an Invertible Neural Network is a neural network that represents an bijective function. With invertibility, it can transform a distribution to another by change of variable formula. Some invertible neural networks decompose the density of data in an auto-regressive manner, and they are called auto-regressive flows \cite{auto1, auto2}. However, here we mainly focus on normalizing flows\cite{RNVP, GLOW2}, which are simpler invertible neural networks that do not assume particular factorization of distribution. 

A normalizing flows usually has carefully designed structures (except recent works such as \cite{iresnet} that do not require special structure) to ensure analytical invertibility and tractable computation of its determinant. With these properties, we can express the log probability of data $\mathbf{x}$ in terms of the log probability of variable $\mathbf{z}$, which has simple distributions, and the log determinant of the flow $f$:
\begin{align} \label{change_variable}
    \log p(\mathbf{x})=\log p(\mathbf{z}) +\log\left|\operatorname{det}\left(\frac{\partial f(x)}{\partial x}\right)\right|.
\end{align}

Assume $p(\mathbf{z})$ is some simple distribution, such as standard Gaussian, we can explicitly compute the likelihood of $x$ and train the flow $f$ by maximizing the likelihood. Likelihood is the most popular training objective for normalzing flows, although other objectives such as adversarial loss or MMD loss can also be used\cite{INN,gan rnvp}.

To ensure invertibility and tractable Jacobian determinant computation, various coupling layers are invented. Coupling layers are the building blocks of normaling flows, as a normalizing flow stacks coupling layers together. Perhaps affine coupling layer introduced in \cite{RNVP} is the most popular one. Given a $D$ dimensional input data $z$ and $d<D$, the input is partitioned into two vectors $\mathbf{z}_{1}=\mathbf{z}_{1:d}$ and $\mathbf{z}_{2}=\mathbf{z}_{d+1:D}$. The output of one affine coupling layer is given by
\begin{align}\label{aff_c}
    \begin{array}{l}{\mathbf{y}_{1}=\mathbf{z}_{1} \odot \exp \left(s_{1}\left(\mathbf{z}_{2}\right)\right)+t_{1}\left(\mathbf{z}_{2}\right)} \\ {\mathbf{y}_{2}=\mathbf{z}_{2} \odot \exp \left(s_{2}\left(\mathbf{y}_{1}\right)\right)+t_{2}\left(\mathbf{y}_{1}\right)}\end{array}
\end{align}
The inverse of this coupling layer is easily computable:
\begin{align}\label{aff_cinv}
    \begin{array}{l}{\mathbf{z}_{2}=\left(\mathbf{y}_{2}-t_{2}\left(\mathbf{y}_{1}\right)\right) \odot \exp \left(-s_{2}\left(\mathbf{y}_{1}\right)\right)} \\ {\mathbf{z}_{1}=\left(\mathbf{y}_{1}-t_{1}\left(\mathbf{z}_{2}\right)\right) \odot \exp \left(-s_{1}\left(\mathbf{z}_{2}\right)\right)}\end{array}
\end{align}
Note that inverting am affine coupling layer does not require the inverse of $s_1, s_2, t_1, t_2$, so they can be arbitrary functions, such as neural networks. 

\subsection{Modeling Inverse Problem with Invertible Neural Networks}

Generally speaking, the inverse problem aims to determine the latent variables given some observations. In particular, the posterior  distribution on the latent variables, conditioned on the observations, needs to be estimated. Just as in other Bayesian inference problems, inverse problems have  previously been handled by variational inference \cite{vi1} or MCMC\cite{mc1}. Very recently, \cite{INN} proposed a novel  idea to analyze inverse problems with invertible neural networks. 

Specifically, suppose $\mathbf{x} \in \mathbb{R}^{D}$ represents the variable that we are interested in, and $\mathbf{y} \in \mathbb{R}^{M}$ is the observation. Suppose we know the forward process $\mathbf{y}=s(\mathbf{x})$. Since the transformation from $\mathbf{x}$ to $\mathbf{y}$ incurs an information loss, the intrinsic dimension of $\mathbf{y}$ is in general smaller than $D$, even if the
nominal dimensions satisfy $M > D$ (think about the case when $y$ is an one-hot label, with intrinsic dimension $1$ but possibly large nominal dimension). Here our task is learning to estimate the conditional probability $p(\mathbf{x}|\mathbf{y})$, given a large amount of training data $\left\{\left(\mathbf{x}_{i}, \mathbf{y}_{i}\right)\right\}_{i=1}^{N}$ where $\mathbf{y}_i's$ come from the known forward process $s$. 

The main idea of \cite{INN} is to introduce dummy random variables $z$ and represent the posterior $p(\mathbf{x}|\mathbf{y})$ by a deterministic function $\mathbf{x}=g(\mathbf{y}, \mathbf{z})$ that transforms the known distribution $p(\mathbf{z})$ to $\mathbf{x}$ space, conditioned on $\mathbf{y}$. Both the forward process $s$ and the backward process $g$ are constructed as a single normalizing flow $f$ with parameter $\theta$. The dummy variable $\mathbf{z} \in \mathbb{R}^{K}$ is drawn from standard Gaussian distribution. Therefore, in the forward direction, we have 
\[
[\mathbf{y}, \mathbf{z}]=f(\mathbf{x} ; \theta)=\left[f_{\mathbf{y}}(\mathbf{x} ; \theta), f_{\mathbf{z}}(\mathbf{x} ; \theta)\right] \quad \text{with} \quad f_{\mathbf{y}}(\mathbf{x} ; \theta) \approx s(\mathbf{x})
\]
and in the backward direction, we have
\[
\mathbf{x}=f^{-1}(\mathbf{y}, \mathbf{z} ; \theta) \quad \text { with } \quad \mathbf{z} \sim p(\mathbf{z})=\mathcal{N}\left(\mathbf{z} ; 0, I_{K}\right)
\]
If the resulting nominal output
dimension $M+K>D$, then we augment $x$ with a zero vector $\mathrm{x}_{0} \in \mathbb{R}^{M+K-D}$. As in \eqref{change_variable} we can express the posterior $p(\mathrm{x} | \mathrm{y})$ as
\[
\log p(\mathbf{x}| \mathbf{y})= \log p(\mathbf{z}) - \log \left|\text{det}\left(\frac{\partial f^{-1}(\mathbf{y}, f_{\mathbf{z}}(\mathbf{x}) ; \theta)}{\partial[\mathbf{y}, \mathbf{z}]}\right)\right|
\]

\cite{INN} trains the model using losses for both directions ending up with a total of $4$ loss terms:
\begin{itemize}
    \item A forward regression loss that learns the forward process $s$  \[\mathcal{L}_{\mathbf{y}}\left(\mathbf{y}, \mathbf{x}\right) = ||\mathbf{y} - f_{\mathbf{y}}\left(\mathbf{x}\right)||_2\]
    \item A loss for the dummy variables that enforces two things: first,
    the generated $\mathbf{z}$ must follow the standard Gaussian distribution and second, $\mathbf{y}$ and $\mathbf{z}$ must be independent. It is implemented by Maximum Mean Discrepancy (MMD).
    \[
    \mathcal{L}_{\mathbf{z}}\left(\mathbf{y}, \mathbf{x}\right)  = \text{MMD}\left( \left[f_{\mathbf{y}}(\mathbf{x} ; \theta), f_{\mathbf{z}}(\mathbf{x} ; \theta)\right], [\mathbf{y}, \mathbf{z}]\right), \quad \mathbf{z} \sim \mathcal{N}\left(\mathbf{z} ; 0, I_{K}\right)
    \]
    \item A backward distribution loss, which ensures that reversing the true dummy variable sampled from prior will result in a  distribution similar to the original data $x$.
    \[
    \mathcal{L}^2_{\mathbf{x}}(x,y)  = \text{MMD}(f^{-1}(\mathbf{y},\mathbf{z}),\mathbf{x}), \quad \mathbf{z} \sim \mathcal{N}\left(\mathbf{z} ; 0, I_{K}\right)
    \]
    \item A backward regression loss to ensure that if the dummy variable obtained by the forward process is slightly perturbed  (sample $\mathbf{\epsilon}$ from a Gaussian distribution with small variance), we can still obtain a good  reconstruction of $\mathbf{x}$.
    \[
    \mathcal{L}^1_{\mathbf{x}}(x,y) = ||x - f^{-1}(\mathbf{y}, f_{\mathbf{z}}(\mathbf{x} ; \theta) + \mathbf{\epsilon})||_2
    \]
    
\end{itemize}

Note that only the first two loss terms are justified by the design of the model. However, the later two losses, which can be seen as a type of cycle consistency, are found to be crucial when implementing the method. They are necessary ad-hoc tricks to stabilize training. Furthermore the authors choose an ad-hoc weighting of the different losses.

\subsubsection{Limitations of Previous Approach}

Despite \cite{INN} successfully use invertible neural networks to analyze inverse problems from both synthetic data and real data, their approach has several major limitations.

\begin{enumerate} \label{limitation}
  \item Most importantly, \cite{INN} heavily relies on MMD loss to train the model, and it is known that MMD  does not easily scale to high-dimensional problems. 
  \item The training objective involves four loss terms, and their relative weights needs to be fine tuned. For example, the author of \cite{INN} propose to gradually increase the weights of backward losses through training, otherwise the model can get stuck. It is not easy to balance multiple loss terms, and we always prefer simple loss function. In addition, the method involves losses in both forward and backward directions, and therefore the gradients have to be evaluated twice, which significantly slows down the training. 
  \item The method does not maximize the posterior $p(\mathbf{x}|\mathbf{y})$ explicitly, instead, it estimates the posterior implicitly. However, it is desirable to explicitly find the most likely $\mathbf{x}$ given observation $\mathbf{y}$.  
\end{enumerate}

\subsection{Conditional Generative Models}
As mentioned in Section \ref{intro}, some attempts have been made to add conditioning in deep generative models. Conditional VAEs\cite{CVAE} modifies the ELBO objective to make each distribution in ELBO (encoder distribution, decoder distribution and prior) conditioned on the given condition. The conditioning can easily implemented by concatenate the condition with inputs. Conditional GANs\cite{CGAN} incorporates conditioning in GANs, and the condition can be concatenated to the input noise or as part of the conditional normalization layers\cite{norm}. Adding conditioning to normal zing flow is also straightforward\cite{CNF}, and we will introduce it as well as compare it with our method in Section \ref{cnf}. 

\section{Our Approach} \label{our}
In this section, we introduce our proposed approach to solve the same inverse problems as in \cite{INN}. Our method can overcome all the limitations described in Section \ref{limitation}.

We train the normalizing flow that models inverse problems by directly maximize the conditional probability $p(\mathbf{x}|\mathbf{y})$. Rather than treating the output of $f_{\mathbf y}(\mathbf x,\theta)$ as $\mathbf y=s(\mathbf x)$ directly, we assume $f_{\mathbf y}(\mathbf x,\theta)$ to be an approximation to $\mathbf y$, and it follows a distribution determined by $\mathbf y$:
\[
f_{\mathbf y}(\mathbf x,\theta) \sim \mathcal{N}(\mathbf y,\epsilon^2 I)
\]
In other words, we can reparametrize $\mathbf{y}$ as

\begin{align}\label{relax}
f_{\mathbf y}(\mathbf x,\theta) = \mathbf{y}+\mathbf{v}, \quad \mathbf{v} \sim \mathcal{N}(0,\epsilon^2 I)
\end{align}

Here $\epsilon$ is relatively small. With the distribution of $f_\mathbf{y}(\mathbf x,\theta)$, we can express the conditional probability  $p(\mathbf{x}|\mathbf{y})$ to be

\begin{align}
    \log p(\mathbf{x}|\mathbf{y}) &= \log p(z) + \log p(f_{\mathbf y}(\mathbf x,\theta) | \mathbf{y}) - \log \left|\operatorname{det}\left(\frac{\partial f^{-1}([\mathbf{y} + \mathbf{v}, \mathbf{z}]; \theta)}{\partial [\mathbf{v},\mathbf{z}]}\right)\right|\notag\\
    & =\log \mathcal{N}(\mathbf{z}; 0, I_K)+\log \mathcal{N}\left( \mathbf{v}; 0, \epsilon^{2} I_{M}\right)+\log \left|\operatorname{det}\left(\frac{\partial f(\mathbf{x}; \theta)}{\partial \mathbf{x}}\right)\right|
\end{align}


The Gaussian likelihood of $\mathbf{v}$ corresponds to the forward regression loss in Section \ref{limitation} (plus an additional term $\frac{1}{2}\log2\pi\epsilon^2$) with weight controlled by the choice of $\epsilon$. As $\epsilon \rightarrow 0$, the distribution of $f_{\mathbf y}(\mathbf x,\theta)$ approaches delta distribution. Although \cite{INN} assumes a deterministic forward process, some noise has to be added to $\mathbf{y}$ when implementing the method.  

Note that in this expression, we have that $\mathbf{z}$ and  $f_{\mathbf{y}}(\mathbf{x} ; \theta)$ are independent, as they are both factorial Gaussian random variables. The log determinant is easily computatable by the design of normalizing flows. Therefore, we can evaluate the exact conditional density explicitly, and perform maximum likelihood estimation to update the parameters of the flow.  

We argue that our method can overcome the limitations of the previous approach. First of all, our method can easily scale to high dimensions, since previous works such as Glow\cite{GLOW2} have shown that  very high dimensional variables can be trained by maximum likelihood. Our method contains only a single loss, which corresponds to the explicit conditional probability. Although one may argue that the likelihood loss still has three terms: two Gaussian likelihood terms and one log determinant term, but they are integrated to form the likelihood, so we do not need to (actually, we should not) tune their weights. Consequently, training using our method is much easier. Our method explicitly maximize the posterior, which could potentially lead to a better solution.

\section{Conditional Normalizing Flow} \label{cnf}

We further notice that our training method provides a way to incorporate conditioning into normalizing flows. Very recently, \cite{CNF} proposes to train normalizing flows with conditioning. The main idea is quite simple. Since we do not need to invert the function $s(\mathbf{x})$ and $t(\mathbf{x})$ in the coupling layer of normalizing flows, they can be arbitrary functions. In \cite{CNF}, the authors simply concatenate the condition $\mathbf{c}$ with $\mathbf{x}$ as the input to coupling layers, so that the affine coupling layer defined in \eqref{aff_c}  becomes 
\[
\begin{array}{l}{\mathbf{y}_{1}=\mathbf{z}_{1} \odot \exp \left(s_{1}\left(\mathbf{z}_{2}, \mathbf{c}\right)\right)+t_{1}\left(\mathbf{z}_{2}, \mathbf{c}\right)} \\ {\mathbf{y}_{2}=\mathbf{z}_{2} \odot \exp \left(s_{2}\left(\mathbf{y}_{1}, \mathbf{c}\right)\right)+t_{2}\left(\mathbf{y}_{1}, \mathbf{c}\right)}\end{array}
\]
where $\left[\mathbf{z}_{1}, \mathbf{z}_{2}\right]$ and $\left[\mathbf{y}_{1}, \mathbf{y}_{2}\right]$ are the splits of the input and output, respectively. Inverting the coupling layer is performed as in \eqref{aff_cinv}. \cite{CNF} test their conditional normalizing flow on the task of label-conditioned generation on MNIST and image coloring. Results show that the proposed model can successfully incorporate the condition in the generation process.

Our method can be easily used for the task of conditional generation. To see this, consider a normalizing flow $f$ with parameter $\theta$ that connects two sets of variables $[\mathbf{x}, \mathbf{x_0}]$ and $[\mathbf{z}, \mathbf{c}]$. Here $\mathbf{x}$ is the data, $\mathbf{c}$ is the condition and $\mathbf{z}$ is the dummy variable. $\mathbf{x_0}$ corresponds to the zero padding to $\mathbf{x}$ in order to match the dimensions. This is not necessary, as we can let $\mathbf{z}$ have smaller dimensionality to ensure that dimensions of $\mathbf{x}$ and $[\mathbf{z}, \mathbf{c}]$ agree, but we include zero padding for generality. In the forward direction , $f([\mathbf{x}, \mathbf{x_0}]) = [f_{\mathbf{z}}([\mathbf{x}, \mathbf{0}]), f_{\mathbf{c}}([\mathbf{x}, \mathbf{0}])]$. We again assume that, as in \eqref{relax},
\begin{align*}
    f_{\mathbf{c}}([\mathbf{x},\mathbf{x_0}]) = \mathbf{c} + \mathbf{v}, \quad \mathbf{v} \sim \mathcal{N}(0,\epsilon^2 I)
\end{align*}
and it is independent of the dummy variable $\mathbf{z}$, which marginally has standard Gaussian distribution. 

Denote $[\mathbf{x}, \mathbf{x_0}]$ as $\hat{\mathbf{x}}$. we can write the conditional probability $p([\mathbf{x}, \mathbf{x_0}]|\mathcal{C})$ as :
\begin{align} \label{likelihood}
    \log p([\mathbf{x}, \mathbf{x_0}]|\mathbf{c}) &=  \log p(\mathbf{z}) + \log p(f_{\mathbf{c}}([\mathbf{x},\mathbf{x_0}])|\mathbf{c} ) + \log \left|\operatorname{det}\left(\frac{\partial f([\mathbf{x}, \mathbf{x_0}]; \theta)}{\partial \hat{\mathbf{x}}}\right)\right|
    \notag\\
    & = \log \mathcal{N}(\mathbf{z}; \mathbf{0}, I_K)+\log \mathcal{N}\left( \mathbf{v}; 0, \epsilon^{2} \mathbf{I}\right)+\log \left|\operatorname{det}\left(\frac{\partial f([\mathbf{x}, \mathbf{x_0}]; \theta)}{\partial \hat{\mathbf{x}}}\right)\right|.
\end{align}
Note that when there is zero padding, the LHS of \eqref{likelihood} is the conditional log likelihood of $[\mathbf{x}, \mathbf{x_0}]$. When $\mathbf{x_0}$ is absolute constant $0$, this is equivalent to the  conditional log density of $\mathbf{x}$. However, in practice, we usually add a small amount of noise to $\mathbf{x_0}$ to avoid degeneracy. In this case, $\mathbf{x_0}$ is actually a Gaussian random variable with mean $0$ and very small standard deviation $\sigma$, and we can write the conditional density of $\mathbf{x}$ as: \begin{equation}
\label{lik_x}
p(\mathbf{x}|\mathbf{c}) = \int p([\mathbf x,\mathbf x_0]|\mathbf{c})d\mathbf x_0.\end{equation} 

We should maximize \eqref{lik_x}, but it is not easy to compute it. Therefore we seek an alternative way to train the flow $f$ by maximizing the joint conditional probability $\log p([\mathbf{x}, \mathbf{x}_0]|\mathbf{c})$. Note that to avoid degeneracy, we have to add a small amount of noise to the zero padding. It can be done by using either independently sampled zero padding $\mathbf {x_0} \sim \mathcal N(\mathbf 0, \sigma^2 I)$ or zero padding with fixed noise in each iteration. Experiments show that these training strategies works very well. After we train the flow model, we can generate samples $\Tilde{x}$ with condition $\mathcal{C}$ by sampling $z$ from the standard Gaussian and $\mathbf{c}$ from $\mathcal{N}\left( \mathcal{C}, \epsilon^{2} \mathbf{I}\right)$ and obtain $\Tilde{x} = f^{-1}([\mathbf{z}, \mathbf{c}])$.

\subsection{Difference from Previous Approaches} \label{difference}
Our conditional normalizing flow shares some similarity with \cite{CNF}, as both use a concatenated input for the coupling layer. However, our method differs from theirs in  that we do not use the condition $\mathbf{c}$ in the forward direction. Instead, we learn a mapping from input $\mathbf{x}$ to a relaxed random variable associated with the condition (and the dummy variable $z$). The conditioning variable can be mapped back to $\mathbf{x}$ in the inverse direction. Doing so allows us to express the conditional probability $\log p([\mathbf{x}, \mathbf{x_0}]|\mathcal{C})$ in a different way than in \cite{CNF}. 

Our method has an additional benefit compared to \cite{CNF} in that our conditional normalizing flow can also be used to infer the condition if necessary. Suppose we are given a data point $\mathbf{x}$, we can obtain an approximation $\mathbf{c} = f_{\mathbf{c}}([\mathbf{x},\mathbf{x}_0])$ with a sampled $\mathbf{x_0}$. Since we enforce the distribution of ${\mathbf{c}}$ to be close to the true condition $\mathcal C$, the obtained $\mathbf{c}$ will recover the true condition with high accuracy. This can be useful in some cases. For example, if we train a conditional normalizing flow to generate MNIST images conditioned on one-hot label, we can use the same network to predict the label of unseen MNIST data simply by feeding the data into $f_{\mathbf{c}}$, and obtain an (approximately) one-hot vector, which can be used to assign label to the data. This is possible for neither the conditional normalizing flow in \cite{CNF}, nor some earlier conditional generation models such as conditional VAEs\cite{CVAE} and conditional GANs\cite{CGAN}. In summary we bleieve our model has the advantage that a single network can be used for both conditional generation and inference.

\section{Experiments}\label{exp}
In this section, we demonstrate the effectiveness of our model for analyzing inverse problems and conditional generation. We only include some preliminary experiments, and more comprehensive empirical evaluation will be left for future work.

\subsection{Toy data: Gaussian Mixtures}
We first apply our method to the Gaussian Mixtures data, which is a toy example for inverse problems used in \cite{INN}. The synthetic dataset contains coordinates of samples $\mathbf{x}$ and their corresponding one-hot labels $\mathbf{y}$ from a standard 8-component Gaussian mixture distribution. The task is to learn the posterior distribution $p(\mathbf{x}|\mathbf{y})$. We adopt the same settings and the same normalizing flow structure as in \cite{INN}. Specifically, we let $z\in\mathbb{R}^2$, and therefore we pad the 2-dimensional $\mathbf{x}$ with $\mathbf{0} \in \mathbb{R}^8$. The normalizing flow has $8$ flow coupling blocks, each containing $3$ fully connected layers with $512$ hidden units, followed by a random but fixed permutation. We choose the standard deviation of $\mathbf{y}$ to be $0.1$, and we add Gaussian noise with standard deviation $0.05$ to the zero padding, which coincides with the noise level in \cite{INN}. We train the model for $400$ iterations with batch size $1600$. We use Adam optimizer with default learning rate $0.001$.

\begin{figure}[htbp]
    \centering
    \includegraphics[width=0.6\textwidth]{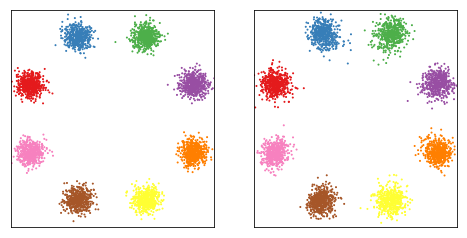}
    \hfill
     \caption{\label{our_GMM}
\textbf{Left}: predicted labels for test data through the forward process. \textbf{Right}: generated samples from the estimated posterior through the backward process.}
\end{figure}

\begin{figure}[htbp]
    \centering
  \centering
    \subfloat{\includegraphics[width=0.3\textwidth]{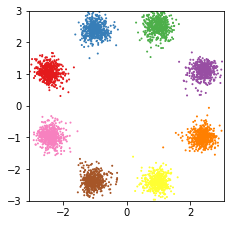}}
    \hfill
    \subfloat{\includegraphics[width=0.3\textwidth]{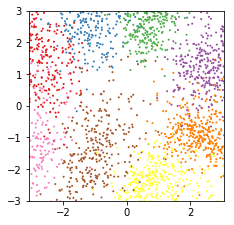}}
    \hfill
    \subfloat{\includegraphics[width=0.3\textwidth]{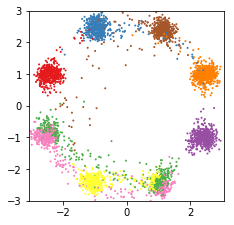}}
  
    \caption{\label{inn_GMM}
    Generated samples from learned posterior using method in \cite{INN}. \textbf{Left}: with full losses and fine tuned weights. \textbf{Middle}: with only forward losses and tuned weights. \textbf{Right}: with full losses and different weights for backward losses. In particular, \cite{INN} linearly increase the weight of backward losses during training, and we change the speed of the weight increase. }
\end{figure}

In Figure \ref{our_GMM}, we show the results of the Gaussian Mixture experiment. We observe that our method can recover the posterior distribution almost exactly. We also compare our results with the results obtained from \cite{INN} in Figure \ref{inn_GMM}. We can see that when trained with all loss terms and carefully tuned weights, the method in \cite{INN} can match the performance of our method. However, their performance are significantly worse when dropping the backward losses (which are not justified by the model) or changing the relative weights of some loss terms. Therefore, our method is advantageous because we can obtain good performance with a single loss and no hyper-parameter tuning.

\subsection{Conditional Generation}
In this section, we evaluate the ability of our conditional normalizing flow by generating class conditional samples of MNIST and Fashion MNIST datasets. Instead of applying the flow on the original image data which are high dimensional, we take the approach of latent flow \cite{GLF} which learns low dimensional latent variables using an auto-decoder, and fits the distribution of the latent variables by a normalizing flow. It is shown that the latent flow model can generate high quality samples with significantly shorter training time compared to plain normalizing flows on the original data. Our normalizing flow builds the connection between distributions of $[\mathbf{u}, \mathbf{0}]$ and $[\mathbf{z}, \Tilde{\mathbf{y}}]$, where $\mathbf{u}$ is the latent variable returned by the deterministic encoder, $\mathbf{z}$ is the noise dummy variable that we assume to follows a standard Gaussian distribution, and $\Tilde{\mathbf{y}}$ is the relaxed random variable corresponding to the one-hot label $\mathbf{y}$. In MNIST example, $\mathbf{z}$ and $\mathbf{\epsilon}$ are 20-dimensional, while the label $\mathbf{y}$ and zero padding $\mathbf{0}$ are 10-dimensional. 

We use the same auto-encoder structure as in \cite{GLF}, and we use the same flow structure as in \cite{GLF}, except that the input and output dimensions increase by $10$ to fit the 10-dimensional one-hat label. Other training details (batch size, optimizer, number of training epochs etc.) also largely follow the original paper. The noise level for $\Tilde{\mathbf{y}}$ is $0.1$, and the noise level for zero padding is $0.05$. 

\begin{figure}[htbp]
  \centering
    \subfloat{\includegraphics[width=0.6\textwidth]{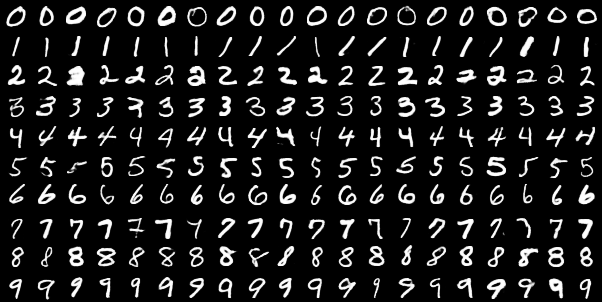}}
    \hfill
    \subfloat{\includegraphics[width=0.6\textwidth]{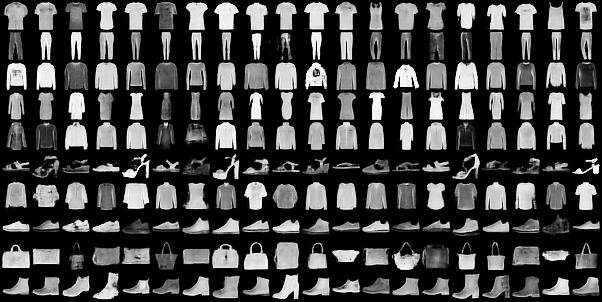}}
    
    \caption{\label{cond_gen}
    Generated samples conditioned on label. Each row contains 20 samples conditioned on a specific one-hot label. \textbf{Top}: MNIST. \textbf{Bottom}: Fashion MNIST.}
\end{figure}

\begin{figure}[htbp]
    \centering
    \includegraphics[width=0.5\textwidth]{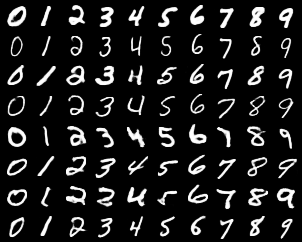}
    \hfill
     \caption{\label{style}
    Each row contains 10 generated images with a fixed dummy variable $\mathbf{z}$ and $10$ different labels $\mathbf{y}$. Note that the dummy variable captures the style of a digit independent from the class label. Styles include thickness of the stroke (for example, row 1 and 2), slanted left or right (row 5 and 6), etc. }
\end{figure}

We present some qualitative results of conditional generation in Figure \ref{cond_gen}. We observe that our model can accurately generate samples with the given conditioning label, and the generated samples are sharp and diverse. To illustrate that the dummy variable $\mathbf{z}$ is independent of $\mathbf{y}$, we generate samples with fixed $\mathbf{z}$ and vary $\mathbf{y}$ in Figure \ref{style}. We observe that the dummy variables successfully encode the style of a digit independent from the condition, as the same style can be shared across different conditions.

In addition, as mentioned in Section \ref{difference}, our conditional normalizing flow can also be used as an inference network for condition variables. In particular, we can run the forward process of the network to classify the images in the test set, which are not used for training. In Table \ref{acc}, we present the classification accuracy of our model. We observe that the normalizing flow gives decent classification accuracy on both datasets.

\begin{table}[htbp]
  \caption{Classification accuracy by running the forward process}
  \label{acc}
  \centering
  \begin{tabular}{lll}
   \toprule
       & MNIST   &  Fashion-MNIST \\
    \midrule
	 accuracy &  0.984  &  $\quad$ 0.901\\
    \bottomrule
  \end{tabular}
\end{table}

\section{Conclusion and Future Works}

In this work, we propose a method to model a conditional probability using invertible neural networks. It can be applied to a variety of problems which involve conditional distributions, such as inverse problems and conditional generation. Preliminary experiments show the effectiveness of our method. We also carefully compare our method with some previous work, and highlight the differences. Applying our method to more complicated tasks, such as solving practical inverse problems will be left for future works.

\end{document}